\documentclass[conference,compsoc]{IEEEtran}
\ifCLASSOPTIONcompsoc
  \usepackage[nocompress]{cite}
\else
  \usepackage{cite}
\fi

\usepackage[utf8]{inputenc} % allow utf-8 input
\usepackage[T1]{fontenc}    % use 8-bit T1 fonts
\usepackage{hyperref}       % hyperlinks
\usepackage{url}            % simple URL typesetting
\usepackage{booktabs}       % professional-quality tables
\usepackage{amsfonts}       % blackboard math symbols
\usepackage{nicefrac}       % compact symbols for 1/2, etc.
\usepackage{microtype}      % microtypography
\usepackage{xcolor}         % colors

\usepackage{amsmath}
\usepackage{amssymb}
\usepackage{graphicx}
\usepackage{multirow}
\usepackage{enumitem}
\usepackage{subcaption}

\usepackage{pifont}
\newcommand{\cmark}{\ding{51}}%
 
\usepackage{algorithm}
\usepackage{algpseudocode}

\newtheorem{definition}{Definition}

\newcommand{\cX}{\mathcal{X}}

\newcommand{\bfone}{\mathbf{1}}
\newcommand{\bfh}{\mathbf{h}}
\newcommand{\bfa}{\mathbf{a}}
\newcommand{\bfx}{\mathbf{x}}
\newcommand{\bfX}{\mathbf{X}}
\newcommand{\bfq}{\mathbf{q}}
\newcommand{\bfQ}{\mathbf{Q}}

\newcommand{\bfZ}{\mathbf{Z}}
\newcommand{\bfW}{\mathbf{W}}
\newcommand{\bfS}{\mathbf{S}}
\newcommand{\bfT}{\mathbf{T}}
\newcommand{\bfU}{\mathbf{U}}

\newcommand{\bbN}{\mathbb{N}}
\newcommand{\bbI}{\mathbb{I}}

\newcommand{\COUNT}{\textsc{Count}}

\begin{document}
\title{Generate-then-Verify:\\ Reconstructing Data from Limited Published Statistics}


% \author{

% \IEEEauthorblockN{Michael Shell}
% \IEEEauthorblockA{School of Electrical and\\Computer Engineering\\
% Georgia Institute of Technology\\
% Atlanta, Georgia 30332--0250\\
% Email: http://www.michaelshell.org/contact.html}

% \and

% \IEEEauthorblockN{Homer Simpson}
% \IEEEauthorblockA{Twentieth Century Fox\\
% Email: homer@thesimpsons.com}

% \and

% \IEEEauthorblockN{James Kirk\\ and Montgomery Scott}
% \IEEEauthorblockA{Starfleet Academy\\
% San Francisco, California 96678-2391\\
% Telephone: (800) 555--1212\\
% Fax: (888) 555--1212}

% }


% compsoc conferences compared to traditional conferences), use this

% \author{\IEEEauthorblockN{Anonymous Authors}
% \IEEEauthorblockA{Paper under review}}
\newcommand*\samethanks[1][\value{footnote}]{\footnotemark[#1]}
\author{
\IEEEauthorblockN{
    Terrance Liu\IEEEauthorrefmark{1}, % 0000-0003-3807-7508
    Eileen Xiao\IEEEauthorrefmark{1}, % 0009-0003-3963-5345
    Adam Smith\IEEEauthorrefmark{2}, % 0000-0001-9393-1127
    Pratiksha Thaker\IEEEauthorrefmark{1}, and % 0000-0001-9977-5081
    Zhiwei Steven Wu\IEEEauthorrefmark{1} % 0000-0002-8125-8227
    \IEEEcompsocitemizethanks{
        \IEEEcompsocthanksitem First two authors contributed equally. Remaining authors are ordered alphabetically.
    }
}
\\
\IEEEauthorblockA{\IEEEauthorrefmark{1}Carnegie Mellon University}
\IEEEauthorblockA{\IEEEauthorrefmark{2}Boston University}}


\maketitle

% \fi
\IEEEpeerreviewmaketitle

\begin{abstract}

We study the problem of reconstructing tabular data from aggregate statistics, in which the attacker aims to identify interesting claims about the sensitive data that can be verified with 100\% certainty given the aggregates. 
Successful attempts in prior work have conducted studies in settings where the set of published statistics is rich enough that entire datasets can be reconstructed with certainty.
In our work, we instead focus on the regime where many possible datasets match the published statistics, making it impossible to reconstruct the entire private dataset perfectly (i.e., when approaches in prior work fail).
We propose the problem of partial data reconstruction, in which the goal of the adversary is to instead output a \textit{subset} of rows and/or columns that are \textit{guaranteed to be correct}. We introduce a novel integer programming approach that first \textbf{generates} a set of claims and then \textbf{verifies} whether each claim holds for all possible datasets consistent with the published aggregates.
We evaluate our approach on the housing-level microdata from the U.S. Decennial Census release, demonstrating that privacy violations can still persist even when information published about such data is relatively sparse.
    
\end{abstract}
\section{Introduction}

% 2010 county lookup: https://www.census.gov/library/reference/code-lists/ansi.2010.html#cou

\begin{table*}[t!]
\centering
\caption{
    Examples of verified, singleton claims (multiplicity $m = 1$)
}

\begin{tabular}{p{1in} p{5in}}
\toprule
\normalsize Block, Tract, &  \multirow{2}{*} {\normalsize  Reconstructed Information} \\
\normalsize County, State & \\
\toprule

% 010030102001008
\normalsize 
1008, 010200, Baldwin, AL  & 
% 010200-1008 & 
\normalsize 
A household with just a single female householder. She owns the home without a mortgage. The householder is white, of Hispanic or Latino origin, and is between 65 and 75 years old. \\

\midrule

% 245102718013027
\normalsize 
3027, 271801, Baltimore City, MD & 
% 271801-3027 & 
\normalsize 
A renting household of size 2. It is a non-family household, and no one in the household is under 18 or over 65 years old. The householder is black, not of Hispanic or Latino origin, and between 25 and 34 years old. \\

\midrule

% 261635645021006
\normalsize 
1006, 564502, Wayne, MI & 
% 564502-1006 & 
\normalsize 
A household of size 4 with a married couple that owns the home with a mortgage. No one in the household is over 65 years old, but there is at least one child under 18 years old. The householder is Black, not of Hispanic or Latino origin, and is between 45 and 54 years old. \\

\midrule

% 320030058281049
\normalsize 
1049, 005828, Clark, NV & 
% 05828-1049 & 
\normalsize 
A married couple household (of unknown size) that does not own the home but also does not pay rent. No one in the household is over 65 years old, but there is at least one child under 18 years old. The householder of Hispanic or Latino origin and between 25 and 34 years old. Their race does not belong to one of the 5 major census race categories. \\

\midrule

% 380539401001087
\normalsize 
1087, 940100, McKenzie, ND & 
% 940100-1087 & 
\normalsize 
A renting household of size 4. There is a cohabiting couple living with at least one child under 18 years old. No one in the household is over 65 years old. The householder is male, American Indian/Alaskan Native, not of Hispanic or Latino origin, and between 15 and 24 years old. \\

\bottomrule
\end{tabular}
\label{tab:example_reconstructions}
\end{table*}

The problem of data privacy lies at the heart of data stewardship.
While many organizations aim to provide data products that maximize utility for downstream users, this goal is at direct odds with protecting the privacy of those who contribute data.
In this paper, we study this problem from the perspective of tabular data reconstruction, in which an adversary is given access only to a set of aggregate statistics about the private dataset. Specifically, we are interested in the setting in which the adversary aims to reconstruct (some portion of) the private dataset with absolute certainty. In other words, we answer the question, \textit{``What must exist in the private dataset according to the published statistics?''}

The aforementioned problem of data stewardship is at the forefront of issues faced by the U.S. Census Bureau, which provides billions of statistics to the public while needing to fulfill a legal mandate to protect the privacy of its respondents \cite{abowd2023confidentiality}. 
As a result, the bureau itself has conducted various studies investigating the vulnerability of the US Decennial Census release to potential reconstruction attacks.
Most recently, for example, Abowd et al.~\cite{abowd20232010} tackle this problem from the lens of guaranteeing correctness (as part of a larger set objectives in their work) and find that by using 34 person-level tables from the 2010 Summary File 1, one can reconstruct the entire data for 70\% of the blocks in the United States with 100\% certainty simply by solving an integer program.
% 6,207,027 total blocks
% (comprised of approximately 97,238,000 person records) 
% Like previous works \cite{TODO}, \cite{abowd20232010} set up data reconstruction as an integer programming problem. However, they augment this standard approach by solving a second optimization problem that verifies whether the solution is unique (and therefore is guaranteed to be correct).

Such alarming results suggest that reconstruction of person-level data using the Decennial Census release is far too easy---the amount of information (statistics) available to the adversary is so rich that reconstruction becomes trivial for the majority of blocks. 
In light of this observation, one might ask whether releasing less descriptive statistics that do not admit a unique IP solution would be sufficient to protect individuals' information.
% One might then ask whether the same is true for more ambiguous datasets in which the statistics provided are relatively less descriptive and do not straightforwardly imply exact solutions---does more ambiguity provide more privacy protection? 
In this work, we therefore study to what extent data reconstruction with 100\% certainty can still occur even in more difficult regimes in which the released statistics are not informative enough (relative to the size of the data domain) for prior approaches (e.g., Abowd et al.~\cite{abowd20232010}) to reconstruct \textit{entire} tabular datasets with absolute certainty.
% \pt{briefly describe what the barrier would be that makes them ineffective?}

\noindent \textbf{Contributions.} We summarize our contributions as the following:
\begin{enumerate}[itemsep=0pt, leftmargin=20pt]
    \item We introduce the problem of \textit{partial} tabular data reconstruction to help better understand the vulnerability of data releases like the Decennial Census: rather than reconstructing the entire dataset with guaranteed correctness, the adversary aims to output \emph{verified} claims about individuals in 
    the data (see, e.g., Figure \ref{fig:example_partial_recon}).
    
    \item We consider claims about the number of rows with specific values in a subset of columns, such as ``in this dataset, there exists exactly one household whose head of the household is a 32-year old, Black woman." In particular, inspired by Cohen and Nissim~\cite{cohen2020towards}, we focus on reconstructing \textit{``singleton claims''}, which are reconstructed attributes that single out exactly one individual in the dataset. 
    
    \item We introduce an 
    integer programming formulation that departs from the approaches of previous work \cite{abowd20232010, dwork2024synthetic, steed2024quantifying} and 
    allows us to tackle this problem. 
    Specifically, given some set of aggregate statistics about the dataset, our method \textbf{(1)} generates a set of candidate claims %(over all combinations of columns) 
    and then \textbf{(2)} verifies whether these claims must be true according to the published statistics. 

    \item We evaluate our approach and that of previous work on the household unit-level data and tables from the Decennial Census release (2010 Summary File 1 (SF1)). We find that the method proposed in Abowd et al.~\cite{abowd20232010} for reconstructing entire blocks is ineffective---in our experiments, not a single block could be reconstructed uniquely.
    In contrast,  our approach reconstructs many individual households with 100\% certainty, demonstrating that partial reconstruction is still feasible, even when full reconstruction is not (Table \ref{tab:example_reconstructions} provides examples of verified claims). 
    
    \item We find that a nontrivial number of households can be reconstructed using some subset of columns that uniquely identifies them (i.e., singles them out). Among the blocks evaluated in our experiments, approximately 40\% contain at least one household that can be singled out by 8 (out of 10 total) columns (Figure \ref{fig:results_unique}), averaging out to one household per block (Table \ref{tab:results_unique}). For 6 columns, the percentage of blocks containing singled out households increases to over 80\% (Figure \ref{fig:results_unique}).
    
\end{enumerate}

\subsection{Additional Related Work}

Real-world examples of privacy risks resulting from aggregate statistical releases have long been well-documented \cite{sweeney1997weaving, narayanan2008robust, abowd2023confidentiality, steed2024quantifying, flaxman2025risk}. As a result, a long line of research, beginning with the seminal work of Dinur and Nissim~\cite{dinur2003revealing}, have both studied reconstruction attacks using public statistical information \cite{dwork2007price, kasiviswanathan2010price, kasiviswanathan2013power, dwork2017exposed} and developed notions for privacy guarantees---namely, Differential Privacy \cite{dwork2006calibrating}.

Mitigating such privacy risks \cite{steed2024quantifying, flaxman2025risk, dick2023confidence, garfinkel2019understanding, abowd20232010, abowd2023confidentiality} remains at the center of issues facing the U.S. Census Bureau, which has addressed such privacy concerns by incorporating Differential Privacy into the 2020 Decennial Census release \cite{Abowd18}. Our work, in part, extends such findings, further demonstrating the risks that individuals face when aggregate statistics derived from them are released freely. While Abowd et al.~\cite{abowd20232010} show that at the person-level, the majority of blocks can be completely reconstructed, their method relies on the Decennial Census release being rich enough so that there can only exist one possible set of individuals that correspond to the released statistics. At the household-level, in which there are far more columns but relatively the same number of statistics, this condition is no longer met---we find that for any given block, there exists many possible solutions (groups of households) that would produce the same set of released statistics. Nevertheless, we devise a method that can partially reconstruct a block with absolute certainty.

Lastly, the notion of singleton claims, which later works like Cohen and Nissim~\cite{cohen2020towards} expand upon, can be traced back to as early as Sweeney~\cite{sweeney1997weaving}, which exposed the susceptibility of uniquely identified individuals to linkage attacks. As a by-product of reconstructing entire datasets (e.g., Abowd et al.~\cite{abowd20232010}), one can single out individuals by identifying the rows that are unique in the reconstruction. We, however, make the observation that complete reconstruction is not strictly necessary for the purpose of singling out. Focusing on \textit{partial} reconstruction, our work demonstrates that individual records can still be reconstructed with certainty and that some individuals can be singled out even by just a subset of the columns in the data domain.

% \pt{I think the thing missing is an explanation that Abowd et al want to reconstruct entire records (which may be infeasible), while we make the observation that you don't need entire records to single out an individual, and partial reconstruction can still be harmful in a singling-out sense}
\section{Preliminaries}\label{sec:prelims}

\begin{figure*}[t!]
    \centering
    \includegraphics[width=0.7\linewidth]{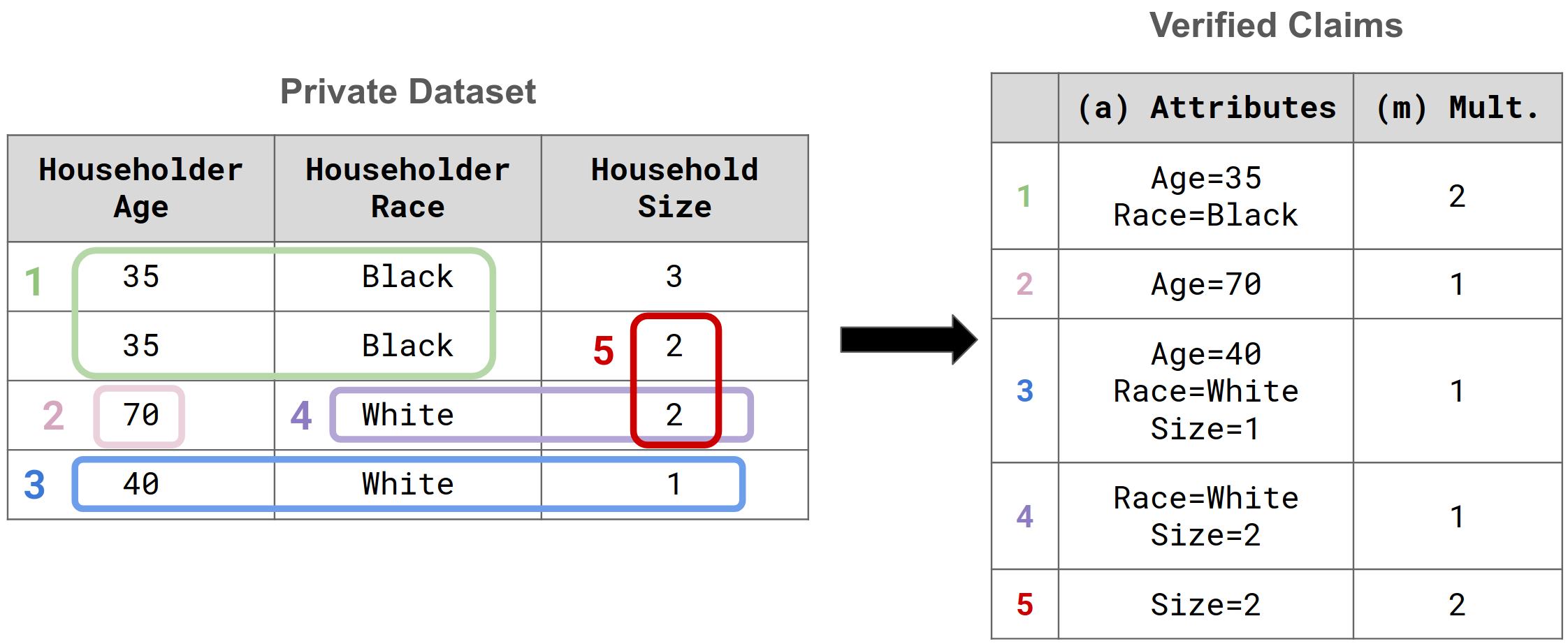}
    \caption{
        We provide a visual diagram of example claims studied in our work. On the left-hand side is the private dataset, where the colored boxes denote various claims $R(a, m)$ that are then enumerated in the table on the right-hand side.
    }
    \label{fig:example_partial_recon}
\end{figure*}

% \subsection{Tabular Data Reconstruction}

In this setting, we have some \textit{dataset} $D$ that is comprised of a multiset of $N$ records from a discrete domain $\cX$. 
Let $Q$ be some set of $n$ queries corresponding to the data domain $\cX$, and let $Q(D) \in \mathbb{R}^n$ be a vector of aggregate statistics on dataset $D$ where each element is a statistic corresponding to a query in $Q$. 
Then, in its most general form, tabular data reconstruction can be set up as a simple constraint satisfaction problem (i.e., find any dataset $D'$ that matches the statistics $Q$),
\begin{align}
    % X = \arg\min_{D'} \ &0 \\
\text{Find } D' \quad \text{s.t.} \ &
Q(D) = Q(D')
%\| Q(D) - Q(D') \|_1 = 0. 
\label{eq:ip_problem}
\end{align}
In our work, we consider statistics in the form of counting queries

\begin{equation}\label{eq:counting_queries}
   q_\phi(D) = \sum_{x\in D} \phi(x),
\end{equation}
where $\phi(x)$ denotes the condition that indicates whether a row $x$ satisfies some property. Thus, $q_\phi(D)$ counts the number of rows $x \in D$ that satisfy that property. Our work focuses on $k$-way marginal queries\footnote{In the typical formulation of $k$-way marginals, $\phi$ checks whether a column is equal to one specific value (e.g., $\textsc{Age} = 10$). Our work considers a more general definition, where the column can take on a set of values (e.g., $\textsc{Age} \in \{10, 20\})$.}, where the $\phi$ indicates whether a set $k$ columns matches some set of values (e.g., $\textsc{Sex}=Male$ and $\textsc{Age} \in \{10, 20\})$. Table \ref{tab:example_table} provides an example of a set of queries tabulated in the U.S. Decennial Census Release.

\begin{table*}[ht!]
    \centering
    \caption{
        We provide an example of a table (Summary File 1: P20) released in the Decennial Census, including the text descriptions of each query contained in the table and the count of households matching that description for some block. In detail, \textbf{Condition $\phi$} denotes what each query checks for (e.g., query 1 checks whether column HHT2 = 1), and indented rows mean that the corresponding query must satisfy the condition \textit{and} all parent conditions above them. For example, query number 10 corresponds to HHT2 = 9, while query number 11 corresponds to HTT2 = 8 \textit{AND} THHLDRAGE = 7, 8, or 9. \textbf{Text Description} describes what each value means. For example, HHT2 = 1 means that the household is a \textit{married couple household with their own children under the age of 18}.
    }
    \begin{tabular}{cllc}
    \toprule
    \textbf{Query No.} & \textbf{Condition $\phi$ (Column = Value)} & \textbf{Text Description} & \textbf{Count} \\
    \midrule
    & & \textit{Married couple household:} & \\
    1 & HHT2 = 1 & \quad With own children under 18 & 6 \\
    2 & HHT2 = 2 & \quad No own children under 18 & 1 \\
    \midrule
    & & \textit{Cohabiting couple household:} & \\
    3 & HHT2 = 3 & \quad With own children under 18 & 0 \\
    4 & HHT2 = 4 & \quad No own children under 18 & 0 \\
    \midrule
    & & \textit{Female householder, no spouse or partner present:} &  \\
    5 & HHT2 = 5 & \quad Living alone & 0 \\
    6 & \quad THHLDRAGE = 7, 8, or 9 & \quad \quad 65 years and over & 0 \\
    7 & HHT2 = 6 & \quad With own children under 18 & 1 \\
    8 & HHT2 = 7 & \quad With relatives, no own children under 18 & 0 \\
    9 & HHT2 = 8 & \quad No relatives present & 0 \\
    \midrule
    & & \textit{Male householder, no spouse or partner present:} &  \\
    10 & HHT2 = 9 & \quad Living alone & 1 \\
    11 & \quad THHLDRAGE = 7, 8, or 9 & \quad \quad 65 years and over & 0 \\
    12 & HHT2 = 10 & \quad With own children under 18 & 0 \\
    13 & HHT2 = 11 & \quad With relatives, no own children under 18 & 1 \\
    14 & HHT2 = 12 & \quad No relatives present & 0 \\
    \bottomrule
    \end{tabular}
    \label{tab:example_table}
\end{table*}

\subsection{Record-level reconstruction.} 

In our work, we focus on \textit{record}-level reconstruction, where the goal is to output claims about sets of rows $x \in \cX$. Suppose there exists $k$ columns in $\cX$ such that we rewrite $\cX = \cX_1 \times \cX_2 \times \ldots \times \cX_k$. Let $\cX'_i = \cX_i \cup \{\bot\}$, where $\bot$ indicates that column $i$ can take on any value in $\cX$. 
A vector $a$ in $\cX' = \cX_1' \times \ldots \times \cX_k'$ specifies a \emph{partial assignment} to the attributes, and we say $x$ matches $a$ if they agree in all coordinates where $a_i \neq \bot$.

%ADAM's NOTE: I think ''attributes'' is confusing, since we are referring to a partial assignment to attributes. (Age is an attribute; 40 is an age...) 
%Then we can describe any $x \in \cX$ by some subset of columns using its corresponding representation in $\cX'= \prod_i^k \cX'_i$ (here, $X_i = \bot$ for any columns $i$ that are omitted). Going forward, we will refer to $a \in \cX'$ as a set of \textit{attributes}, which denotes both some (sub)set of columns and the values that they take on. 

Using this notation, we define $R(a, m)$ to be the (reconstruction) \textbf{claim} that there exist \textbf{exactly} $m \in \{0, 1, \dots, N\}$ rows (e.g., $m=2$ in Figure\ref{fig:example_partial_recon}; claim 1) that match $a \in \cX'$ (e.g., $a$ describes a 35-year old, Black householder in Figure\ref{fig:example_partial_recon}; claim 1).
We can then define a \textit{singleton} claim as some claim $R(a,m)$ where $m=1$.

Let $\COUNT(a, D) : \cX' \times \cX^N \rightarrow \bbN \cup \{0\}$ be the number of rows in $D$ that match $a$. Then we say a claim $R(a, m)$ is correct for some dataset $D$ if $\COUNT(a, D) = m$. 

Finally, we define \textit{verified} claims as the following:
\begin{definition}[Verified Claim]\label{def:verified_claim}
Given some set of summary statistics $Q(D)$, we say that a claim $R(a,m)$ is \emph{verified} with respect to $Q(D)$ if and only if
\begin{equation*}
    \COUNT(a, D') = m
\end{equation*}
for all datasets $D'$ such that $Q(D)=Q(D')$.
\end{definition}
In other words, a claim is verified (i.e., guaranteed to be correct) if it must be correct for any dataset $D'$ where $Q(D)=Q(D')$.

\subsection{Guaranteeing the correctness of claims.}

At a high level, to verify the correctness of any claim $R(a, m)$, one can ask the question: is it possible to construct a synthetic dataset $D'$ that matches the published statistics, even when the multiplicity of number of rows with attributes $a$ does not equal $m$? If such a dataset $D'$ does not exist, then $R(a, m)$ must be correct. Concretely then, we check claims by again solving Problem \ref{eq:ip_problem} but with the added constraint that $\COUNT(a, D) \ne m$:
\begin{alignat}{2}
    % X' = \arg\min_{D'} &\ 0 \\
\text{Find} \quad D' \quad   
    \text{s.t.} \quad & 
    Q(D) = Q(D') %\| Q(D) - Q(D') \|_1 &= 0 
    \mbox{ and } \COUNT(a, D) \ne m. \label{eq:ip_check} 
\end{alignat}
Note that in this formulation, we can make use of all statistics (queries $Q$ defined over all columns in $\cX$) available to us, even when verifying claims that are defined over only some subset of columns in $\cX$ (attributes $a$ where $a_i = \bot$ for some columns $i$).

\subsection{Generate-then-Verify}

At a high level, our approach can be broken down into two integer programming steps:
\begin{enumerate}[itemsep=1pt, leftmargin=20pt]
    \item \textbf{Generate:} We generate a list of claims $R(a, m)$ that we then verify in step 2. Specifically, we solve Problem \ref{eq:ip_problem} $K=100$ times.\footnote{In Gurobi, we can simply set the solver to output up to $K$ solutions that satisfy the constraints.} For each generated synthetic dataset $D'$, we identify all claims $R(a, m)$ (i.e., all possible combinations of attributes $a$ and the corresponding multiplicities $m$ in $D'$). We then take the intersection of the $K$ sets of claims to use as our final list.\footnote{The intersection contains \textit{all} claims that are plausible based on the aggregate statistics $Q(D)$. If a claim does not belong in the intersection, then there exists some feasible reconstruction $D'$ consistent with $Q(D)$ that refutes the claim, meaning that we cannot be certain that the claim is correct for $D$. We explain this filtering logic further in \ref{sec:generating_candidates}.}
    
    \item \textbf{Verify:} For each claim $R(a, m)$, we check if Problem \ref{eq:ip_check} is feasible via integer programming. If no solutions can be found, then we conclude that $R(a, m)$ must be correct.
\end{enumerate}
We defer to Section \ref{sec:ip_setup} the details of how we encode the input for the integer programming solver.
\section{Empirical Evaluation}\label{sec:experiments}

\subsection{Setup}

\subsubsection{Dataset} In our experiments, we use the 2010 Privacy-Protected Microdata File, a synthetic dataset, statistically similar to the private 2010 Decennial Census microdata, that is generated and released by the U.S. Census Bureau. As the private 2010 Census microdata are not public, we treat the Privacy-Protected Microdata as the ground truth during evaluation. 
The PPMF (and Summary File 1, from which the PPMF is derived from) contains data for every housing unit in the United States. Each row of the PPMF represents one synthetic household response from the 2010 Decennial Census. There are 10 columns in total described by block-level tables (listed in Appendix \ref{appx:experimental}), in contrast to the simpler, person-level data studied in prior work \cite{abowd20232010, dick2023confidence} that only contains 4 columns.

\begin{figure}[ht!]
    \centering
    \includegraphics[width=0.7\columnwidth]{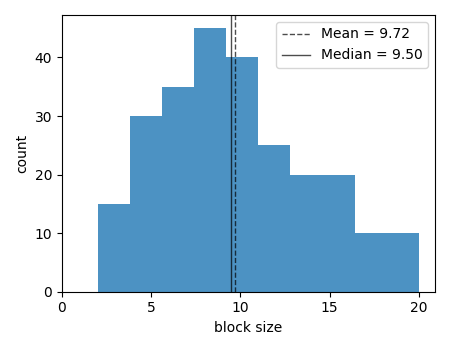}
    \caption{
        We plot the distribution of block sizes for blocks whose size equals the median block size in the state. The minimum block size is 2 and the maximum is 20. 
    }
    \label{fig:state_median_block_sizes}
\end{figure}

From each U.S. state (50 in total), we select blocks in the following ways:
\begin{itemize}
    \item For each state, we calculate the median block size (Figure \ref{fig:state_median_block_sizes}) and randomly select 5 blocks of that size.
    \item We calculate the median block size ($N=10$) of the country and select 5 blocks of that size from each state.
\end{itemize}
Crucially, unlike for the census release of the person-level data studied in Abowd et al.~\cite{abowd20232010}, \textbf{no blocks we evaluate on can be fully reconstructed with 100\% certainty}---when solving Problem \ref{eq:ip_problem}, we found at least 2 different solutions $D'$ for every block, meaning each block $D$ is not uniquely identifiable by the released statistics $Q$.

\subsubsection{Statistics} In addition to the Privacy-Protected Microdata File, the U.S. Census Bureau releases aggregate statistics of features listed above, calculated from their private microdata, in the form of data tables called Summary File 1 (SF1). Each of these tables are released for every block and includes counts for the number of people corresponding to certain feature values defined by the table. As noted previously, these statistics correspond to $k$-way marginal queries, where $k\le4$ columns\footnote{Using the notation presented in Section \ref{sec:ip_setup}, $k \le 4$ is equivalent to saying that $r_{\text{max}}=4$.} for SF1. In total, we have 24 partial sets of $k$-way marginals (see Table \ref{tab:number_of_way}).
\begin{table}[h]
    \centering
    \begin{tabular}{r|cccc|c}
        \toprule
         $k$ & 1 & 2 & 3 & 4 & total \\
        \midrule
         \# $k$-way marginal sets & 4 & 10 & 8 & 2 & 24 \\
         \bottomrule
    \end{tabular}
    \caption{Number of sets of $k$-way marginals per value $k$.}
    \label{tab:number_of_way}
\end{table}

We provide an example of a table from this release in Table \ref{tab:example_table}. Here, query number 11 counts the number of households owned by a female householder who is over the age of 65 and lives alone. As suggested by Table \ref{tab:example_table}, the statistics released by the Census Bureau only partial cover the $k$-way histogram for any set of $k$ attributes. For example, Table \ref{tab:example_table} bins together the values 7, 8, and 9 for column THHLDRAGE and does not tabulate over instances where THHLDRAGE takes on values 1-6.

Utilizing all tables (listed in Appendix \ref{appx:experimental}) tabulated at the block-level, we have $|Q|=621$ queries as inputs to our integer programming approach.

\begin{figure*}[t!]
    \centering
    % state median
    \begin{subfigure}{0.3\textwidth}
        \centering
        \includegraphics[width=\linewidth]{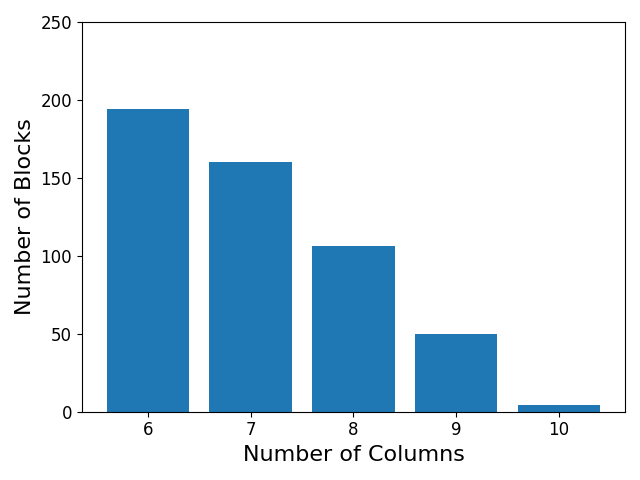}
        \caption{$\ge 1$ Verified Singleton Claims}
        \label{fig:results_unique_a}
    \end{subfigure}
    \hfill
    \begin{subfigure}{0.3\textwidth}
        \centering
        \includegraphics[width=\linewidth]{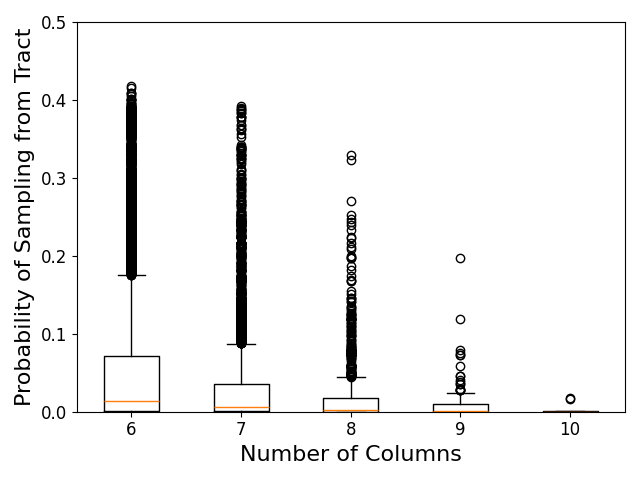}
        \caption{Tract-Level Distribution}
        \label{fig:results_unique_b}
    \end{subfigure}
    \hfill
    \begin{subfigure}{0.3\textwidth}
        \centering
        \includegraphics[width=\linewidth]{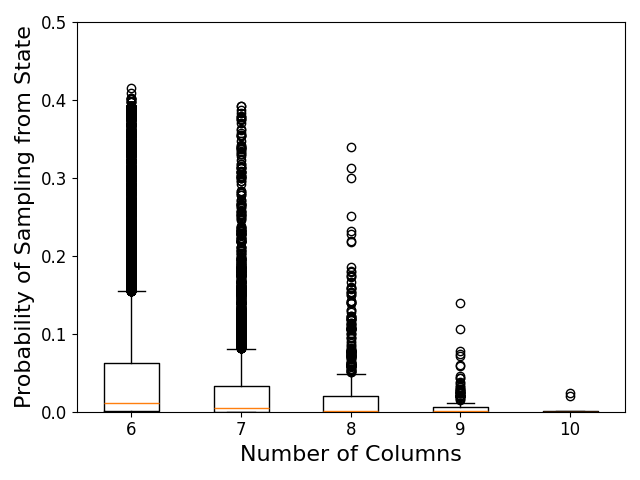}
        \caption{State-Level Distribution}
        \label{fig:results_unique_c}
    \end{subfigure}
    % country median
    \begin{subfigure}{0.3\textwidth}
        \centering
        \includegraphics[width=\linewidth]{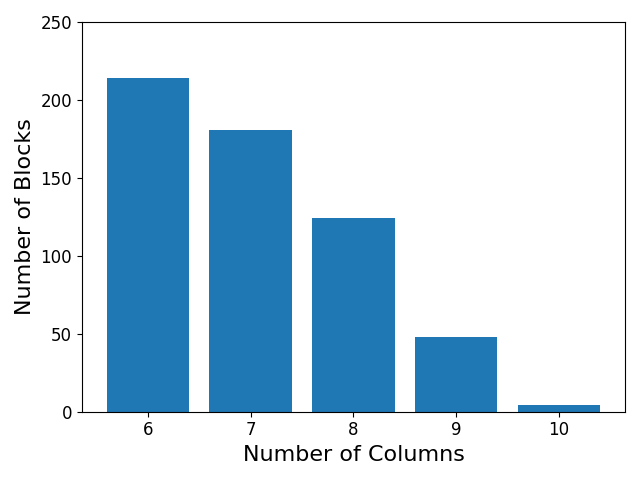}
        \caption{$\ge 1$ Verified Singleton Claims}
        \label{fig:results_unique_d}
    \end{subfigure}
    \hfill
    \begin{subfigure}{0.3\textwidth}
        \centering
        \includegraphics[width=\linewidth]{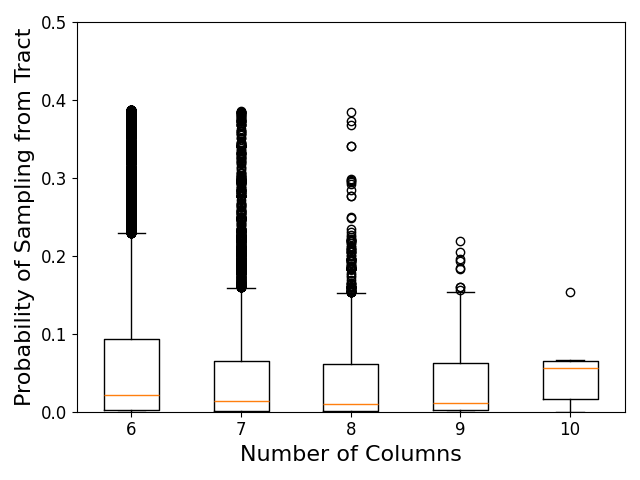}
        \caption{Tract-Level Distribution}
        \label{fig:results_unique_e}
    \end{subfigure}
    \hfill
    \begin{subfigure}{0.3\textwidth}
        \centering
        \includegraphics[width=\linewidth]{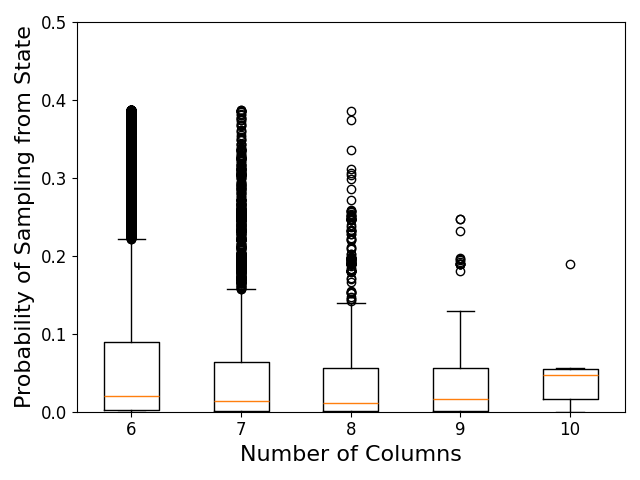}
        \caption{State-Level Distribution}
        \label{fig:results_unique_f}
    \end{subfigure}
    
    \caption{
        We present results for verified \textbf{singleton} claims collected from experiments on 5 blocks selected from each state (250 total on each row).
        \textbf{Top row:} 5 blocks whose size are equal the median block size of the respective state are selected.
        \textbf{Bottom row:} 5 blocks whose size are equal the median block size of the country (i.e., 10 households). 
        \textbf{a \& d:} The number of blocks (out of 50) for which we can reconstruct at least one singleton claim about $k$ columns (x-axis).
        \textbf{b \& e and c \& f:}: Box and whisker plots of the probabilities that each verified claim would also be true in a set of $N$ households randomly sampled from the (b, e) tract and (c, f) state-level distributions. The orange line within each box indicates the median probability. The ends of the box indicate the first and third quartiles, and the whiskers end at the furthest point within 1.5 times the interquartile range. All points beyond the whiskers are outliers. Lower probabilities denote more ``surprising'' claims.
    }
    \label{fig:results_unique}
\end{figure*}

\begin{table*}[t!]
    \centering
    \caption{
        For each number of columns, we report total and average number of households that are represented among the verified \textbf{singleton} claims.
        We tabulate the verified claims over 500 total blocks: (top two rows) 5 blocks from each state whose size is equal to the median block size in the state and (bottom two rows) 5 blocks from each state whose size is equal the country median block size (i.e., 10 households).
        $n$ is the total and average number of households over all 250 blocks.
    }
    \begin{tabular}{llc|ccccc}
        \toprule
        & & & \multicolumn{5}{c}{\# of households identified by } \\
        & & & \multicolumn{5}{c}{verified claims w/ $k$ columns} \\
        & & \# households & $k$=6 & 7 & 8 & 9 & 10 \\
        \midrule
        \multirow{2}{*}{State Median} & Total & 2500 & 659 & 471 & 254 & 97 & 7 \\
        & Avg. per block & 10.00 & 2.64 & 1.88 & 1.02 & 0.39 & 0.03 \\
        \midrule
        \multirow{2}{*}{Country Median} & Total & 2430 & 669 & 437 & 230 & 71 & 6 \\
        & Avg. per block & 9.72 & 2.88 & 1.88 & 0.99 & 0.31 & 0.03 \\
        \bottomrule
    \end{tabular}
    \label{tab:results_unique}
\end{table*}

\subsection{Baseline reconstruction rates}\label{sec:baseline}

While we contend that finding any records that can be reconstructed with $100\%$ confidence is already interesting, we would like to further provide context for our results by providing some baseline measure for how likely a block corroborates some claim. To do so, we calculate the probability of each verified claim being correct in a block of size $N$ that is randomly sampled from the tract or state that the block is located in.

Let us assume that records in a block are drawn from some prior distribution $P$. Then the multiplicity $m$ of some candidate record $x$ appearing in a block $D$ of size $N$ follows the binomial distribution,
\begin{align}\label{eq:baseline}
    P(\COUNT(x, D)=m) = \binom{N}{m}p^m(1-p)^{N-m},
\end{align}
where $p = P(x)$ is the probability of a single record $x$ being drawn from the prior $P$.

In typical settings in which an adversary has no prior information about the block of interest, $p$ is simply the uniform distribution (i.e., $P(x) = \prod_{j=1}^k \frac{1}{|\cX|} = \prod_{j=1}^k \frac{1}{|\bfx^{(j)}|}$, where $\bfx$ is the one-hot encoded representation of $x$ with columns $\{c_j\}_{j=1}^k$). However, this comparison is uninteresting since $P(\COUNT(x, D)=m)$ is close to $0$ in such cases.

In our evaluation, we instead construct a setting in which we assume that the prior distribution of the \textit{tract} and \textit{state} that some block $D$ belongs to is known (similar to baselines considered in Dick et al.~\cite{dick2023confidence}). Let $D_{\text{tract}}$ and $D_{\text{state}}$ be the set of records in the tract and state. Then, we can express $p$ as 
\begin{align*}
    P(x) = \frac{\COUNT(x, \tilde{D})}{|\tilde{D}|}
\end{align*}
for $\tilde{D} = D_{\text{tract}}$ and $\tilde{D} = D_{\text{state}}$, respectively, and use Equation \ref{eq:baseline} to calculate the baseline probability for any given candidate claim $x$.

\section{Results}

We now present our empirical results for verified \textit{singleton} claims.\footnote{Figures and tables for all claims, regardless of multiplicity, can be found in Appendix \ref{appx:additional_results}.} For conciseness, we report results only for claims that cover $k\ge6$ columns since the claims containing more attributes are relatively more interesting.

We also note that there exist claims that can be ``read'' directly off the tables themselves. For example, a table reporting 2 households with White householders already tells us that the claim $R(\textrm{Race=White}, 2)$ must be correct. As mentioned in Section \ref{sec:experiments}, however, the marginal statistics capture at most, $k=4$ columns. Thus, none of the claims reported in this section (i.e., with $k\ge6$) are among this set of ``trivial'' claims.

\subsection{Main findings}\label{sec:results_main}

We present our main results in Figure \ref{fig:results_unique} and Table \ref{tab:results_unique}. In both the figure and table, we split the 500 blocks into two sets: one for blocks whose size equals the national median and one for those whose size equals the respective state median. Interestingly, the results do not differ much across the two sets, suggesting that reconstruction rates do not depend heavily on the size of the blocks we evaluated on.

In Figure \ref{fig:results_unique_a} and \ref{fig:results_unique_d}, we report the number of blocks (y-axis) for which we can reconstruct some set of $k$ columns (x-axis) for \textit{at least} one household. We find that while reconstruction (with 100\% certainty) of all 10 columns is often not possible, we can still partially reconstruct singletons from most blocks. For example, we verify at least one singleton claim with $k=8$ columns in approximately 40\% of the blocks and for $k=6$, we can verify at least one claim in 80\% of them.

In Figures \ref{fig:results_unique_b}, \ref{fig:results_unique_c}, \ref{fig:results_unique_e}, and \ref{fig:results_unique_f}, we evaluate the baseline probabilities (Section \ref{sec:baseline}; Equation \ref{eq:baseline}) for all claims verified by our approach to understand how ``surprising'' they are, given some prior information about the state and tract demographics. Interestingly, the distribution of probabilities is similar for both tract and state-level priors, suggesting using the tract-level prior is no more informative than the state-level one. We find that in general, these baseline probabilities are quite low. In almost cases (except verified claims with $k=10$; Figures \ref{fig:results_unique_e} and \ref{fig:results_unique_f}), the median probability is under 2\% (and often is very close to 0\%). The 75th and 90th percentiles are under 10\% and 25\% respectively, and even among outliers, the maximum baseline probability never exceeds 50\%. As stated previously, we argue that verifying any claim with \textit{100\% certainty} is already interesting and significant. However, these results help demonstrate that if someone were to make guesses about households based on prior information about the tract or state, it is highly unlikely that these guesses would include the claims that our approach outputs.

Finally, in Table \ref{tab:results_unique} we report the number of unique households that we reconstruct, given some number of columns $k$. Here, instead of counting the total number of verified claims, we total up the number of unique households covered by the claims.\footnote{For example, in Figure \ref{fig:example_partial_recon}, the total multiplicity of claims 2 and 4 is two. However, only one household is represented among these claims (row 3 on the left-hand table). Thus, Table \ref{tab:results_unique} groups the verified claims by the number of columns (in $a$) and reports the number of households (in the left-hand table) that are represented in the claims.}
Again, despite the difficulty of reconstructing all $k=10$ columns of households, we find that a non-trivial fraction (over 10\%) of households are uniquely identifiable by some claim that describes $k=8$ columns. This proportion increases to over a quarter when considering claims that describe $k=6$ columns.

\subsection{Ablation: removing single count queries}

\begin{table}[ht!]
    \centering
    \caption{
        We report how the total number of households that are represented among the verified \textbf{singleton} claims changes when different sets of queries are removed as input to our approach. For example, the second row corresponds to removing queries that evaluate to 1 ($q(D) = 1$). We evaluate on blocks (250 in total) whose size is equal to the country median (i.e., $N=10$ for all blocks).
    }
    \begin{tabular}{l|c|ccccc}
    \toprule
    
    & & \multicolumn{5}{c}{\# of households identified by} \\
    & \% queries & \multicolumn{5}{c}{verified claims w/ $k$ columns} \\
    input queries & removed & $k$=6 & 7 & 8 & 9 & 10 \\
    \midrule
    $Q$                           & 0\% & 669 & 437 & 230 & 71 & 6 \\
    $Q \setminus \{q(D) = 1\}$    & 3.40\% & 249 & 132 & 59 & 18 & 0 \\
    $Q \setminus \{q(D) = 0\}$    & 90.88\% & 460 & 246 & 85 & 7 & 0 \\
    $Q \setminus \{q(D) = 0, 1\}$ & 94.28\% & 47 & 20 & 9 & 0 & 0 \\
    \bottomrule
    \end{tabular}
    \label{tab:remove_counts}
\end{table}
\begin{figure*}[t!]
    \centering
    \includegraphics[width=1.0\linewidth]{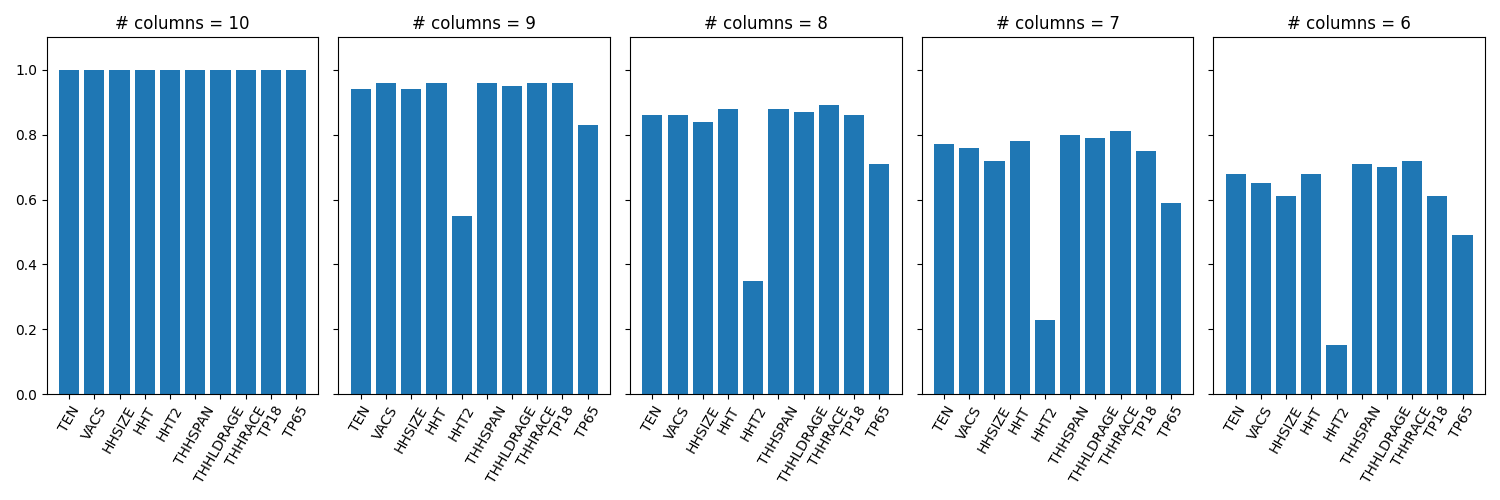}
    \caption{
        For each number of columns $k$, we plot the proportion of verified singleton claims that contain each column.
    }
    \label{fig:claim_columns}
\end{figure*}
\begin{table*}[ht!]
    \centering
    \caption{
        For each number of columns $k$, we list the 5 most common combinations of columns among the verified claims. In addition, we report what percentage of claims with $k$ columns each combination makes up. A checkmark ($\cmark$) indicates that the column is included. For example, 44.54\% of claims comprised of $k=9$ columns omit the column HTT2, while 16.59\% omit the column TP65.
        %
        % Note that $k=10$ is excluded since 100\% of such claims will always include all (10) columns. 
    }
    \begin{tabular}{c|c|c|c|c|c|c|c|c|c|c|c}
        \toprule
        \# columns & TEN & VACS & HHSIZE & HHT & HHT2 & THHSPAN & THHLDRAGE & THHRACE & TP18 & TP65 & \% \\
        \midrule
        10 & \cmark & \cmark & \cmark & \cmark & \cmark & \cmark & \cmark & \cmark & \cmark & \cmark & 100\% \\
        \midrule
        \multirow{5}{*}{9}
& \cmark & \cmark & \cmark & \cmark & & \cmark & \cmark & \cmark & \cmark & \cmark & 44.54\% \\
& \cmark & \cmark & \cmark & \cmark & \cmark & \cmark & \cmark & \cmark & \cmark & & 16.59\% \\
& \cmark & \cmark & & \cmark & \cmark & \cmark & \cmark & \cmark & \cmark & \cmark & 6.11\% \\
& & \cmark & \cmark & \cmark & \cmark & \cmark & \cmark & \cmark & \cmark & \cmark & 6.11\% \\
& \cmark & \cmark & \cmark & \cmark & \cmark & \cmark & & \cmark & \cmark & \cmark & 4.8\% \\
        \midrule
        \multirow{5}{*}{8}
& \cmark & \cmark & \cmark & \cmark & & \cmark & \cmark & \cmark & \cmark & & 12.61\% \\
& \cmark & \cmark & & \cmark & & \cmark & \cmark & \cmark & \cmark & \cmark & 8.01\% \\
& \cmark & & \cmark & \cmark & & \cmark & \cmark & \cmark & \cmark & \cmark & 7.58\% \\
& \cmark & \cmark & \cmark & \cmark & & \cmark & \cmark & \cmark & & \cmark & 7.03\% \\
& \cmark & \cmark & \cmark & \cmark & & \cmark & & \cmark & \cmark & \cmark & 6.82\% \\
        \midrule
        \multirow{5}{*}{7}
& \cmark & \cmark & \cmark & \cmark & & \cmark & \cmark & \cmark & & & 4.68\% \\
& \cmark & \cmark & & \cmark & & \cmark & \cmark & \cmark & \cmark & & 4.37\% \\
& \cmark & & \cmark & \cmark & & \cmark & \cmark & \cmark & \cmark & & 3.81\% \\
& \cmark & \cmark & \cmark & & & \cmark & \cmark & \cmark & \cmark & & 3.13\% \\
& & \cmark & \cmark & \cmark & & \cmark & \cmark & \cmark & \cmark & & 3.07\% \\
        \midrule
        \multirow{5}{*}{6}
& \cmark & \cmark & & \cmark & & \cmark & \cmark & \cmark & & & 3.21\% \\
& \cmark & & \cmark & \cmark & & \cmark & \cmark & \cmark & & & 2.31\% \\
& \cmark & & & \cmark & & \cmark & \cmark & \cmark & \cmark & & 2.03\% \\
& \cmark & \cmark & \cmark & & & \cmark & \cmark & \cmark & & & 1.9\% \\
& & \cmark & \cmark & \cmark & & \cmark & \cmark & \cmark & & & 1.83\% \\
        \bottomrule
    \end{tabular}
    \label{tab:claim_columns}
\end{table*}

In Section~\ref{sec:results_main}, we present results for verified singleton claims that cannot be read directly off the input tables (i.e., number of columns $k > 4$). We note however that in some cases, it might be possible for humans to manually find additional claims without too much difficulty by combining single count queries (queries $q(D) = 1$). To give a toy example, suppose we have the following statistics:
\begin{enumerate}
    \item $\sum \bbI\{A=0, B=0\} = 1$
    \item $\sum \bbI\{B=0, C=0\} =1 $
    \item $\sum \bbI\{B=0\} = 1$
\end{enumerate}
Queries 1 and 2 tell us that there is exactly one row with columns $A=0$ and $B=0$ and one row with $B=0$ and $C=0$. Because query 3 tells us that there exactly one row with $B=0$, we know that query 1 and 2 are describing the same row. Thus, one can look at this set of queries and deduce an additional claim that there is a singleton with the attributes $A=0$, $B=0$, and $C=0$.

Therefore, to further eliminate the possibility of including ``easy'' claims in our results, we simulate a setting where queries that evaluate to 1 are removed from our integer programming approach. Evaluating only on the country median-sized blocks so the block size in our ablation study is fixed, we report in Table \ref{tab:remove_counts} how the number of households that can be uniquely identified changes when the single count queries are removed. We find that although the number of singled out households decreases significantly,\footnote{On average, 3.4\% of queries are removed for each block. However, because 90.88\% of queries evaluate to 0, the single count queries account for almost 37.28\% of nonzero queries counts. As a result, it is unsurprising that the number of verified singleton claims decreases by such a large amount.} a nontrivial number are still uniquely identifiable. For example, approximately 10\% (249 out of 2500) households can still be singled out by $k=6$ columns.

To stress test our approach, we also report in Table \ref{tab:remove_counts} the number of uniquely identifiable households when we remove queries that (a) evaluate to 0 and (b) evaluate to 0 or 1. Unsurprisingly, many of the households are no longer identifiable, especially in the case where queries that evaluate to 0 or 1 are removed. Still, we show that some privacy risks persist, given that a nonzero number of households are singled out by $k=6$ to $8$ columns.

\subsection{Analysis of reconstructed columns}

Finally, in Figure \ref{fig:claim_columns} and Table \ref{tab:claim_columns}, we take a closer look at what columns make up the verified claims outputted by our approach. As shown in Figure \ref{fig:claim_columns}, there generally is an even distribution of columns represented in the verified claims. However, the column HHT2 (detailed household type) is most often omitted, followed by TP65 (presence of someone over 65 years). Examining the most common combinations of $k$ columns reconstructed by approach, we observe similar patterns in Table \ref{tab:claim_columns}. For example, for each number of columns $k$, the most common set of columns does not include HHT2. In fact, HHT2 does not appear at all among the top five most common combinations for $k \le 8$. Similarly, TP65 does not appear in the top for $k \le 7$.

\section{Integer Programming Details}\label{sec:ip_implementation}

In this section, we describe the exact details of our integer programming approach, including how we set up and solve Problems \ref{eq:ip_problem} and \ref{eq:ip_check}. 

\subsection{Setup}\label{sec:ip_setup}

\subsubsection{One-hot encoded records} Unlike prior work \cite{abowd20232010, dwork2024synthetic, steed2024quantifying} which represents datasets as histograms over $\cX$, our proposed integer programming optimization problem relies on one-hot encoded representations of $\cX$. Specifically, let $k$ be the number of columns, which we denote as columns $\{c_j\}_{j=1}^k$, in the domain. Given that all columns in $\cX$ are discrete, we represent records in $\cX$ as one-hot encoded vectors $\bfx = ( \bfx^{(1)} \ldots \bfx^{(k)} )$, where each $\bfx^{(j)}$ encodes the column $c_j$. Thus, we have rows $\bfx \in \{0, 1\}^d$ where $d=\sum_{j=1}^k |\bfx^{(j)}|$ and $k = \sum_{i=1}^d \bfx_i$. Finally, we let the matrix $\bfX \in \{0, 1\}^{N \times d}$ denote a one-hot encoded dataset with $N$ rows.

\subsubsection{Query functions} In this setting, we consider statistical queries (Equation \ref{eq:counting_queries}) in the form of marginal queries where the predicate function $\phi$ is an indicator function for whether some set of columns takes on some set of values (note that $\phi$ is equivalent to what we call attributes $a$ in Section \ref{sec:prelims}). For example, one can ask the marginal query about the columns \textsc{Sex} and \textsc{Race}: ``How many people are (1) \textsc{Female} and (2) \textsc{White} or \textsc{Black}?" We note that one can break down any predicate $\phi$ into a set of {sub-predicates}, where each sub-predicate corresponds to one unique column pertaining to $\phi$.
Concretely, given some column $c$ and target values $V$, we denote the \textit{sub-predicate} function as
\begin{equation*}
    \phi_{c, V}(x) = \bbI\{x_c \in V\},
\end{equation*}
where $x_c$ is the value that $x$ takes on for column $c$. Then, any predicate can be rewritten as the product of its sub-predicates (e.g. in the above example, $\phi$ can be written as the product of $\phi_{\textsc{Sex}, \{\textsc{Female}\}}$ and $\phi_{\textsc{Race}, \{\textsc{White, Black}\}}$). 

Given the one-hot encoded representation $\bfx$, $\phi_{c, V}$ can also be rewritten as a
vector $\bfq \in \{0, 1\}^d$ that takes on the value $1$ for indices in $\bfx$ corresponding column $c$ and values $v \in V$ (and $0$ otherwise). In this case, we can then rewrite the sub-predicate function as $\bfx \bfq^T$. Likewise, any predicate with $r$ sub-predicates $\bfq_1, \bfq_{2}, \dots, \bfq_r$ can be rewritten as a matrix $\bfQ = \begin{pmatrix} \bfq_1 & \bfq_{2} & \dots & \bfq_r \end{pmatrix}^T \in \{0, 1\}^{r \times d}$. Then $\phi$ can be written as $\bbI \{ \bfx \bfQ^T = \bfone_r \}$ where $\bfone_r$ is a row vector of ones with length $r$. Finally, a statistical query $q_\phi$ can be written as
\begin{equation}\label{eq:onehot_query}
    q_\phi(x) = \sum_{j=1}^N \bbI \{ ( \bfX \bfQ^T )_j = \bfone_r \}.
\end{equation}
In other words, we check whether each row in $\bfX$ evaluates to $\bfone_r$.\footnote{We note that a simpler alternative to checking $(\bfQ \bfX^T )_j = \bfone_r$ is to check whether the row product is equal to $1$ (i.e. $\prod_k (\bfQ \bfX^T )_{jk} = 1$). However, our integer programming solver (Gurobi) does not support this operation.}

\subsubsection{Evaluating multiple queries} In our setting, the set of queries $Q$ contain queries that can differ in the number of sub-predicates (i.e., columns that are being asked about). For instance, using the above example data domain, one query may ask about the column \textsc{Sex} while another may ask about both \textsc{Sex} and \textsc{Race}. To handle such cases, given some set of queries $Q$, we let $r_\text{max}$ be the maximum number of sub-predicates for queries in $Q$. Then, in cases where some query predicate is comprised of $r < r_\text{max}$ sub-predicates, we can pad its matrix representation $\bfQ$ with rows corresponding to dummy sub-predicates $\bfq_\text{pad} = \bfone_d$. In this way, Equation \ref{eq:onehot_query} still holds (with $\bfone_r$ being replaced by $\bfone_{r_{\text{max}}}$).

Given that now the matrix representation for all queries in $Q$ have the same shape, we can represent $Q$ as a single $3$-dimensional tensor $\bfQ^{(n)} \in \{0, 1\}^{n \times r_{\text{max}} \times d}$. Then, we can calculate the statistics for all queries in $Q$ by evaluating the product $\bfZ = \bfQ^{(n)} \bfX^T \in \{0, 1\}^{n \times r_{\text{max}} \times N}$, where the $i$-th query answer is
\begin{equation}\label{eq:onehot_query_set}
    Q(X)_i = \sum_{k=1}^N \bbI \{ \bfZ[i,:,k] = \bfone_{r_{\text{max}}} \}.
\end{equation}
Note that we can interpret the vector $\bfZ[i, :, k]$ as a boolean vector that checks whether record $k$ satisfies each of the $r_{\text{max}}$ sub-predicates for query $i$. For ease of notation, we will assume going forward that $r$ refers to $r_{\text{max}}$.

\subsection{Generating Synthetic Data Using Aggregate Statistics (Problem \ref{eq:ip_problem})}

We first describe how we set up the integer programming optimization problem for Problem \ref{eq:ip_problem}---namely, how we represent the constraint $Q(D) = Q(D')$. Using the notation in Section \ref{sec:ip_setup}, we wish to find some synthetic dataset $X$ (whose one-hot representation we denote as $\bfX$) such that $Q(D) = Q(X)$.

As suggested above, we first evaluate whether each record $\bfX_k$ satisfies all $r$ sub-predicates for each query $q_i$ (i.e., $\bfZ[i,:,k] = \bfone_r$). To do so, we want to add a helper binary variable $\bfW \in \{0, 1\}^{n \times N}$ such that
\begin{equation*}
    \bfW[i, k] = \bbI \{ \bfZ[i,:,k] = \bfone_r \}.
\end{equation*}
To enforce this relationship, we add the following constraints,
\begin{align}
    \bfW[i, k] \leq& \bfZ[i, j, k], \quad \forall j \in \{ 1, 2, \ldots, r\} \label{eq:query_constraint_1} \\
    \bfW[i, k] \geq& \sum_{j=1}^{r} \bfZ[i, j, k] - (r-1), \label{eq:query_constraint_2}
\end{align}
so that $\bfW[i, k]$ evaluates query $q_i$ for record $X_k$. Then, we add the constraints
\begin{align}
    Q(D)_i = \sum_{k=1}^{N} \bfW[i,k], \quad \forall i \in \{ 1, 2, \ldots, m\}
\end{align}
to ensure that the aggregate count corresponding to query $i$ on the private dataset $D$ match that on $\bfX$.

\textit{Explanation.} Suppose $\bfW[i, k] = 1$. Given that $\bfZ$ is binary, Equation \ref{eq:query_constraint_1} is true if and only if $\bfZ[i, j, k] = 1$ for all $j$, thereby giving us $\bbI \{ \bfZ[i,:,k] = \bfone_r \} = 1$. Moreover, Equation \ref{eq:query_constraint_2} is not violated since we have that
\begin{align*}
    1 &\ge \sum_{j=1}^{r} \bfZ[i, j, k] - (r-1) \\
      &\ge r - (r - 1) \quad \text{(} \bfZ \text{ is binary)} \\
      &= 1
\end{align*}
Similarly, if $\bfW[i, k] = 0$, then by Equation \ref{eq:query_constraint_2},
\begin{align*}
    0 &\ge \sum_{j=1}^{r} \bfZ[i, j, k] - (r - 1) \\ \implies
    r - 1 &\ge \sum_{j=1}^{r} \bfZ[i, j, k],
\end{align*}
which, because $\bfZ$ is binary, can hold if and only if there exists some $j$ such that $\bfZ[i, j, k] = 0$, meaning that $\bbI \{ \bfZ[i,:,k] = \bfone_r \} = 0$. In this case, Equation \ref{eq:query_constraint_1} is not violated since $0 \le \bfZ[i, j, k]$ (again, because $\bfZ$ can only take on the values $0$ and $1$).

\subsection{Verifying Claims (Problem \ref{eq:ip_check})}

We now discuss the constraints for Problem \ref{eq:ip_check}: verifying whether some claim must be correct in the private dataset $D$ according to the released statistics $Q(D)$.

In this setting, we have a claim $R(a, m)$ that is composed of attributes and claimed multiplicity of that record, $m$. Suppose attributes $a$ are defined over some subset of columns indexed by the set $A$ (i.e, the columns $\{c_j\}_{j \in A}$). Then we can define attributes $a$ as a vector $\bfa = ( \bfa^{(1)} \ldots \bfa^{(k)} )$, where $\bfa^{(j)}$ is a one hot encoded representation of the attribute for column $j$ if all zeros otherwise.

In order to check whether $R(\bfa, m)$ is verifiable  according to $Q(D)$, we stipulate in the optimization problem that the number of times $\bfa$ appears in $\bfX$ cannot equal $m$. If a feasible solution does not exist, then we can conclude that $R(\bfa', m)$ must be correct.

\subsubsection{Constraints (part 1)} We first define a constant $M \gg N$ (used for ensuring other constraints are held) and let $A^{oh}$ correspond to the list of indices that columns $\{c_j\}_{j \in A}$ correspond to in $\bfa$.

Next, let us introduce the binary variable $\bfT \in \{0, 1\}^{N \times d}$, where
\begin{align*}
    \bfT_{ij} =
    \begin{cases} 
    \bf1\{ \bfX_{ij} = \bfa_j \}, & \text{if } j \in A_{oh} \\ 
    1, & \text{otherwise}.
    \end{cases}
\end{align*}
In other words, it indicates whether each column in $\cX$ matches the corresponding column value in $\bfa$. In addition, we introduce $\bfU \in \{0, 1\}^{N \times d}$, which is a helper variable used to set $\bfT$ properly in the constraints. 

Now, we add constraints with $\bfT$ and $\bfU$. Let $v = |A^{(oh)}|$ be the number of (one-hot) indices we need to check. For each index $j \in A^{(oh)}$, we add the constraints,
\begin{align}
\bfX_{i,j} - v \leq& M(1-\bfT_{i,j}) \label{eq:constraint_check_1_1} \\
v - \bfX_{i,j} \leq& M(1-\bfT_{i,j}) \label{eq:constraint_check_1_2}\\
\bfX_{i,j} - v \geq&  1 - M\bfT_{i,j} - M(1-\bfT_{i,j})\bfU_{i,j} \label{eq:constraint_check_1_3} \\
v - \bfX_{i,j} \geq& 1 - M\bfT_{i,j} - M(1-\bfT_{i,j})(1-\bfU_{i,j}). \label{eq:constraint_check_1_4}
\end{align}

% \textit{Explanation.} Consider the case when $\bfT_{i,j} = 1$. Then from constraints \ref{eq:constraint_check_1_1} and \ref{eq:constraint_check_1_2}, we have that $\bfX_{i,j} \leq v$ and $\bfX_{i,j} \geq v$, which implies that $\bfX_{i,j} = v$. Thus, we know that the indicator for the feature value in the one-hot encoding of the candidate is equal to its corresponding value in row $i$ of $\bfX$. Constraints \ref{eq:constraint_check_1_3} and \ref{eq:constraint_check_1_4} give $\bfX_{i,j} - v \geq 1-M$ and $v - \bfX_{i,j} \geq 1-M$. 
% Given that $M$ is a large constant and that $v$ and $\bfX_{i,j}$ are both in the domain $\{0, 1\}$, these constraints are also met.

% Now consider the case when $\bfT_{i,j} = 0$. From constraints \ref{eq:constraint_check_1_3} and \ref{eq:constraint_check_1_4}, we have that $\bfX_{i,j} - v \geq  1 - M\bfU_{i,j}$ and $v - \bfX_{i,j} \geq  1 - M(1-\bfU_{i,j})$. Then, for $\bfU_{i,j} = 0$, we have that 
% $\bfX_{i,j} - v \geq  1$ and $v - \bfX_{i,j} \geq  1 - M$, which implies $1 \leq \bfX_{i,j} - v \leq M - 1$, enforcing that $\bfX_{i,j} \neq v$ when $\bfT_{i,j} = 0$. Similarly, for 
% $\bfU_{i,j} = 1$, we have $\bfX_{i,j} - v \geq  1$ and $v - \bfX_{i,j} \geq 1 - M$ which yields $1 \leq \bfX_{i,j} - v \leq M - 1$ and enforces $\bfX_{i,j} \neq v$ when $\bfT_{i,j} = 0$.

\textit{Explanation.} Consider the case when $\mathbf{T}_{i,j} = 1$. 
From constraints~\ref{eq:constraint_check_1_1} and~\ref{eq:constraint_check_1_2}, we have:
\[
\mathbf{X}_{i,j} \leq v \quad \text{and} \quad \mathbf{X}_{i,j} \geq v 
\quad \Rightarrow \quad \mathbf{X}_{i,j} = v.
\]
Thus, the indicator for the feature value in the one-hot encoding of the candidate 
is equal to its corresponding value in row $i$ of $\mathbf{X}$.

From constraints~\ref{eq:constraint_check_1_3} and~\ref{eq:constraint_check_1_4}, we also have:
\[
\mathbf{X}_{i,j} - v \geq 1 - M 
\quad \text{and} \quad 
\mathbf{X}_{i,j} - v \leq M - 1.
\]
Given that $M$ is a large constant and that $v$ and $\bfX_{i,j}$ are both in the domain $\{0, 1\}$, these constraints are also met.

Now consider the case when $\mathbf{T}_{i,j} = 0$.
From constraints~\ref{eq:constraint_check_1_1} and~\ref{eq:constraint_check_1_2}, we have:
\[
\mathbf{X}_{i,j} - v \leq M \quad \text{and} \quad \mathbf{X}_{i,j} - v \geq - M
\]
Given that $M$ is large, these constraints are always satisfied.

From constraints~\ref{eq:constraint_check_1_3} and~\ref{eq:constraint_check_1_4}, we obtain:
\[
\begin{aligned}
\mathbf{X}_{i,j} - v &\geq 1 - M \mathbf{U}_{i,j}, \\
v - \mathbf{X}_{i,j} &\geq 1 - M (1 - \mathbf{U}_{i,j}).
\end{aligned}
\]

\medskip

\noindent\textit{Case 1:} $\mathbf{U}_{i,j} = 0$
\[
\begin{aligned}
\mathbf{X}_{i,j} - v &\geq 1, \\
v - \mathbf{X}_{i,j} &\geq 1 - M 
\quad \Rightarrow \quad 
1 \leq \mathbf{X}_{i,j} - v \leq M - 1,
\end{aligned}
\]
which enforces $\mathbf{X}_{i,j} \neq v$ (since $\mathbf{X}_{i,j} - v \neq 0$).

\medskip

\noindent\textit{Case 2:} $\mathbf{U}_{i,j} = 1$
\[
\begin{aligned}
\mathbf{X}_{i,j} - v &\geq 1 - M, \\
v - \mathbf{X}_{i,j} &\geq 1
\quad \Rightarrow \quad 
1 \leq v - \mathbf{X}_{i,j} \leq M - 1,
\end{aligned}
\]
which also enforces $\mathbf{X}_{i,j} \neq v$.

%%%%

\subsubsection{Constraints (part 2)} Next, we add the binary variable $\bfS \in \{0, 1\}^{N}$, which is an indicator that checks whether each row $\bfX_i$ matches on the attributes $\bfa$. If an entire row $\bfX_i$ matches the attributes, then the entire corresponding row of $\bfT_i$ should be equal to
1. This can be enforced with the following constraints:
\begin{align}
\bfS_i \leq& \bfT_{i, j} \quad \forall j \in A^{(oh)} \label{eq:constraint_check_2_1} \\
\bfS_i \geq& \sum_{j \in A^{(oh)}} \bfT_{i, j} - (v - 1)\label{eq:constraint_check_2_2} 
\end{align}

\textit{Explanation.} When $\bfS_i = 1$, all values of $\bfT_i$ must be equal to 1 from constraint \ref{eq:constraint_check_2_1}. When $\bfS_r = 0$, we have that at least one of value in row $r$ of $\bfT$ must not be equal to 1 from 
constraint \ref{eq:constraint_check_2_2}.

%%%%

\subsubsection{Constraints (part 3)} With $\bfS$ indicating which rows $\bfX$ match attributes $\bfa$, we now check whether the claimed multiplicity $m$ is correct by summing $\bfS$ and checking if a there exists some dataset $\bfX$ s.t. $\sum_{i=0}^{N-1} \bfS_i \ne m$. If the solver is unable to find a solution $\bfX$ under these constraints, then we conclude that dataset matching the statistics $Q(D)$ cannot exist without having exactly $m$ rows that match attributes $\bfa$.\footnote{
While it is not the focus of our work, we would like to point out that a similar integer programming problem can be set up to confirm that a candidate at multiplicity $m$ \textbf{cannot} exist by replacing constraints \ref{eq:constraint_check_3_1} and \ref{eq:constraint_check_3_2} so that they instead ensure $\sum_{i=0}^{N-1} S_i = m$.
If the solver cannot find a solution where exactly $m$ rows match $\bfa$, then we conclude that candidate \textit{cannot} exist at that multiplicity in the dataset.
}

Let $Y$ be a scalar binary helper variable. Then we add the constraints,
\begin{align}
    \sum_{i=0}^{N-1} \bfS_i - m &\leq MY - 1 \label{eq:constraint_check_3_1} \\
    \sum_{i=0}^{N-1} \bfS_i - m &\geq 1 - M(Y - 1) \label{eq:constraint_check_3_2} 
\end{align}

\textit{Explanation.} Suppose $Y=0$. Then we have that
\[
\sum_{i=0}^{N-1} \mathbf{S}_i - m \leq -1 \quad \text{and} \quad
\sum_{i=0}^{N-1} \mathbf{S}_i - m \geq 1 - M,
\]
which implies that
\[
1 \leq m - \sum_{i=0}^{N-1} \mathbf{S}_i \leq M - 1.
\]
Thus, we enforce that $m \neq \sum_{i=0}^{N-1} \mathbf{S}_i$.

Similarly, suppose $Y = 1$. Then
\begin{align*}
\sum_{i=0}^{N-1} \mathbf{S}_i - m \leq M - 1\ \quad \text{and} \quad
\sum_{i=0}^{N-1} \mathbf{S}_i - m \geq 1,
\end{align*}
which implies that
\[
1 \leq \sum_{i=0}^{N-1} \mathbf{S}_i - m \leq M - 1
\]
Thus, we again enforce that $m \neq \sum_{i=0}^{N-1} \mathbf{S}_i$.

\subsection{Additional implementation details}

\subsubsection{Generating Unique Datasets}

As stated previously, we set the integer programming solver to output up to $K$ solutions that we then use to generate claims. However, in the one-hot encoded representation of datasets, two datasets with the same set of records that are ordered differently will be considered two unique solutions. To encourage unique solutions, we use a (fixed) vector $\bfh \in \bbN^d$ of randomly generated integers as a hash function, where the hash value for any one-hot encoded record $\bfx$ is $\bfh^T \bfx$. Then, for every $i \in \{1, 2, \ldots, N-1\}$, we add the constraint,
\begin{align}
    \bfh^T \bfX_i \geq \bfh^T \bfX_{i-1}
\end{align}
so that any solution $\cX$ outputted must have its records ordered by their hash values. While this approach is imperfect because, theoretically, different records in $\cX$ may map to the same hash value, we found it to be, in practice, a simple and computationally-efficient approach to filtering out duplicate solutions.

\subsubsection{Paring down candidate claims}\label{sec:generating_candidates}

Suppose our goal is to find all reconstruction claims $R$ that \textit{must exist} in $D$ according to $Q(D)$. Let us denote $U(D)$ as the set of all claims that are correct with respect to $D$. Then any claim $R$ is correct if $R \in U(D)$.

Using this notation, our goal of verifying claims to find all claims $R$ such that $\forall D' \in \cX^N$ s.t. $Q(D') = Q(D)$, $R \in U(D')$. In other words, the only way we can be absolutely certain that some claim $R$ is correct according to $Q(D)$ is for it be correct for all datasets $D'$ where $Q(D') = Q(D)$. 

To generate candidate claims, one can simply generate a single synthetic dataset $D'$ by solving Problem \ref{eq:ip_problem} and taking all unique claims $R(a, m)$ consistent with $D'$ (i.e., $\COUNT(a, D') = m$). Furthermore, to narrow down the set of candidates to check, one can generate many synthetic datasets $X_i$ and take their intersection $\bigcap_{i}^K U(X_i)$. If there exists some $R \in U(X_i)$ such that $R \notin \bigcap_{i-1}^K U(X_i)$, then $R$ violates the above condition that $R \in U(D')$ for any $D'$ that matches the released statistics $Q(D)$.

Finally, we note that adjusting $K$ (i.e., the number of synthetic datasets to output in the \textbf{generate} step) allows one to trade-off computational resources between the two steps. Generating more synthetic datasets (i.e., decreasing the size of the intersection) will decrease the number of candidates that need to be verified.
\section{Conclusion}

In conclusion, our work introduces the problem of partial tabular data reconstruction and proposes an integer programming approach that reconstructs individual records with guaranteed correctness. Evaluating on the household-level microdata and tables from the U.S. Decennial Census, we demonstrate that one can still (partially) reconstruct individual households with certainty, even when many possible blocks may satisfy the published statistics. We note that one limitation of using integer programming in our approach is that evaluating on larger datasets (more rows or columns) or sets of statistics may induce computational costs far more demanding than those required for our experiments. Nevertheless, our experiments show that for releases like the decennial census, in which the average dataset (i.e., block) is relatively small, reconstruction is very much possible while being computationally inexpensive. Overall, we contend that our initial work on partial reconstruction represents just the tip of the iceberg in terms of communicating the privacy risks that come with releasing aggregate information. We hope that our work inspires future research to build upon such notions of partial reconstruction (e.g., extending our approach to other data domains or using our approach of singling out households as part of larger, more systematic study on the dangers of linkage attacks).

\bibliographystyle{IEEEtran}
\bibliography{main}

\appendix

\begin{figure*}[t!]
    \centering
    % state median
    \begin{subfigure}{0.3\textwidth}
        \centering
        \includegraphics[width=\linewidth]{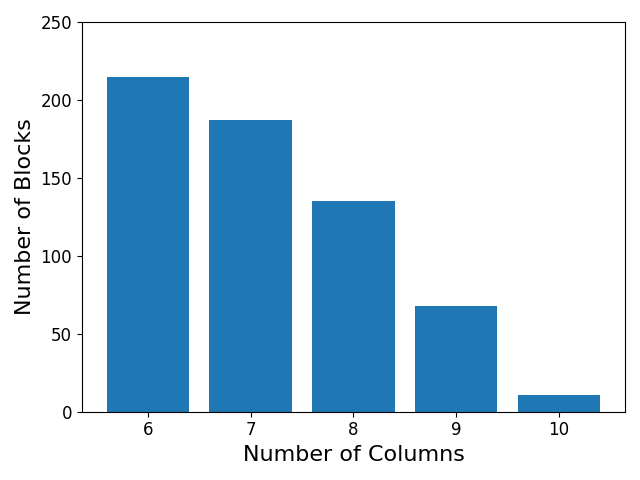}
        \caption{$\ge 1$ Verified Singleton Claims}
        \label{fig:results_a}
    \end{subfigure}
    \hfill
    \begin{subfigure}{0.3\textwidth}
        \centering
        \includegraphics[width=\linewidth]{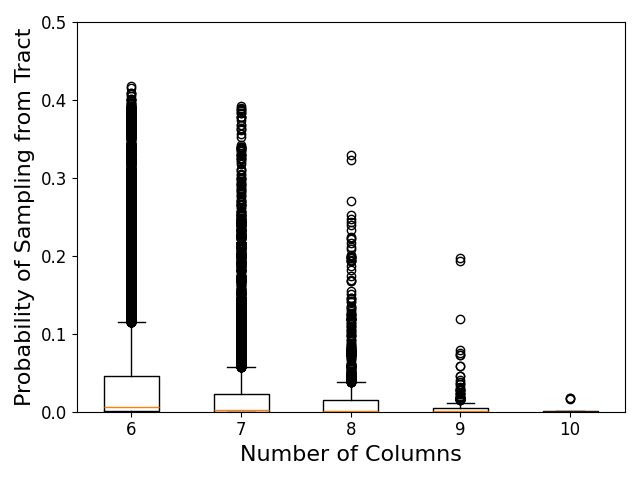}
        \caption{Tract-Level Distribution}
        \label{fig:results_b}
    \end{subfigure}
    \hfill
    \begin{subfigure}{0.3\textwidth}
        \centering
        \includegraphics[width=\linewidth]{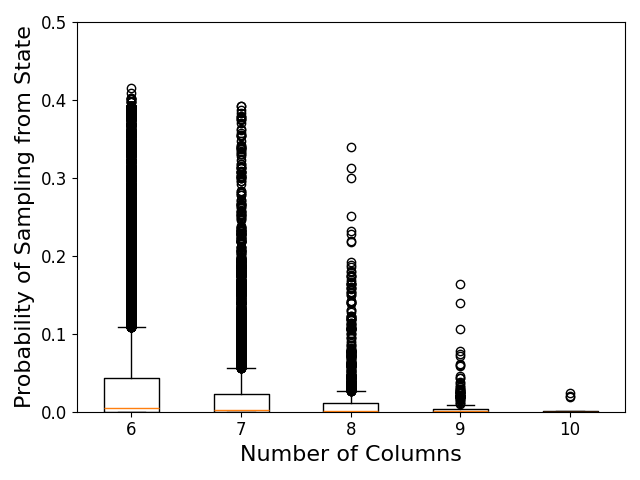}
        \caption{State-Level Distribution}
        \label{fig:results_c}
    \end{subfigure}
    % country median
    \begin{subfigure}{0.3\textwidth}
        \centering
        \includegraphics[width=\linewidth]{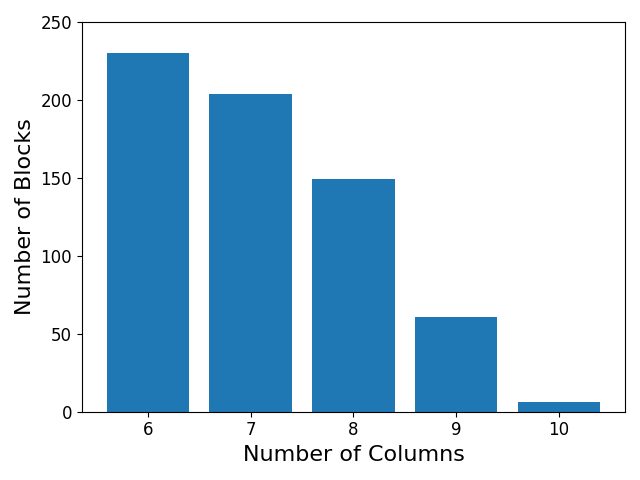}
        \caption{$\ge 1$ Verified Singleton Claims}
        \label{fig:results_d}
    \end{subfigure}
    \hfill
    \begin{subfigure}{0.3\textwidth}
        \centering
        \includegraphics[width=\linewidth]{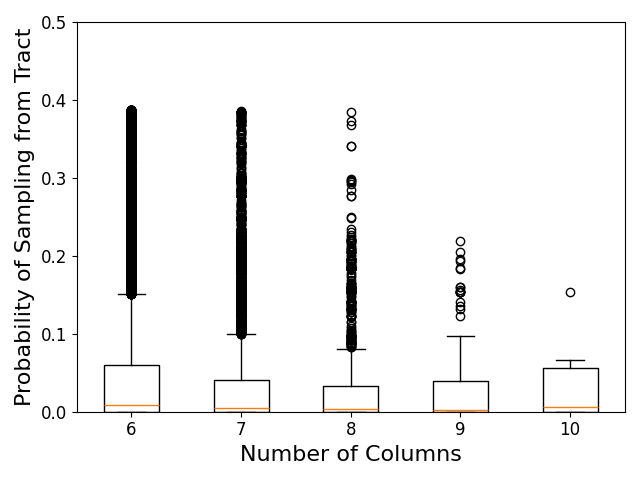}
        \caption{Tract-Level Distribution}
        \label{fig:results_e}
    \end{subfigure}
    \hfill
    \begin{subfigure}{0.3\textwidth}
        \centering
        \includegraphics[width=\linewidth]{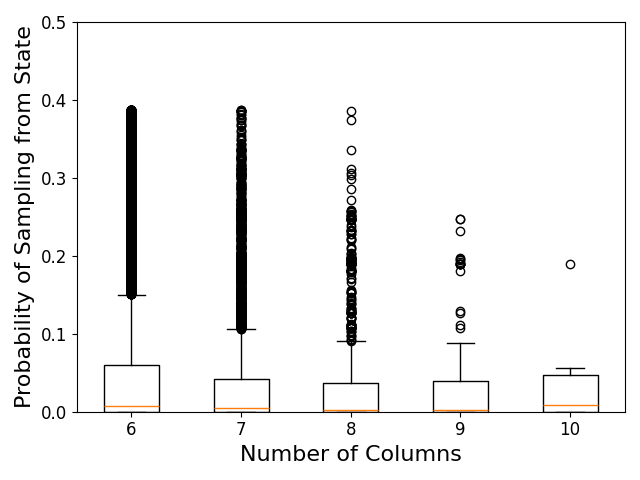}
        \caption{State-Level Distribution}
        \label{fig:results_f}
    \end{subfigure}
    
    \caption{
        We present presents results for \textbf{all} verified claims collected from experiments on 5 blocks selected from each state (250 total on each row).
        \textbf{Top row:} 5 blocks whose size are equal the median block size of the respective state are selected.
        \textbf{Bottom row:} 5 blocks whose size are equal the median block size of the country (i.e., 10 households). 
        \textbf{a \& d:} The number of blocks (out of 50) for which we can reconstruct at least one singleton claim about $k$ columns (x-axis).
        \textbf{b \& e and c \& f:}: Box and whisker plots of the probabilities that each verified claim would also be true in a set of $N$ households randomly sampled from the (b, e) tract and (c, f) state-level distributions. The orange line within each box indicates the median probability. The ends of the box indicate the first and third quartiles, and the whiskers end at the furthest point within 1.5 times the interquartile range. All points beyond the whiskers are outliers. Lower probabilities denote more ``surprising'' claims.
    }
    \label{fig:results}
\end{figure*}

\begin{table*}[t!]
    \centering
    \caption{
        For each number of columns, we report total and average number of households that are represented among the \textbf{all} verified claims.
        We tabulate the verified claims over 500 total blocks: (top two rows) 5 blocks from each state whose size is equal to the median block size in the state and (bottom two rows) 5 blocks from each state whose size is equal the country median block size (i.e., 10 households).
        $n$ is the total and average number of households over all 250 blocks.
    }
    \begin{tabular}{llc|ccccc}
        \toprule
        & & & \multicolumn{5}{c}{\# of households identified by } \\
        & & & \multicolumn{5}{c}{verified claims w/ $k$ columns} \\
        & & \# households & $k$=6 & 7 & 8 & 9 & 10 \\
        \midrule
        \multirow{2}{*}{State Median} & Total & 2500 & 1696 & 1220 & 642 & 257 & 53 \\
        & Avg. per block & 10.00 & 6.78 & 4.88 & 2.57 & 1.03 & 0.21 \\
        \midrule
        \multirow{2}{*}{Country Median} & Total & 2430 & 1809 & 1262 & 691 & 216 & 35 \\
        & Avg. per block & 9.72 & 7.80 & 5.44 & 2.98 & 0.93 & 0.15 \\
        \bottomrule
    \end{tabular}
    \label{tab:results}
\end{table*}

\subsection{Additional experimental details}

\section{Additional Experimental Information}\label{appx:experimental}

\noindent\textbf{Dataset.} We list and describe the 10 columns described by the block-level tables below.

\begin{itemize}[noitemsep, leftmargin=20pt]
  \renewcommand{\labelitemi}{}

\item \textbf{Tenure (TEN)}: One of 4 tenancy statuses: owned with mortgage, owned free and clear, rented, or occupied without payment of rent

\item \textbf{Vacancy status (VACS)}: Not vacant, or one of 7 vacancy statuses: for rent, rented but not occupied, for sale, sold and not occupied, for seasonal or occasional use, for migrant workers, or other.

\item \textbf{Household size (HHSIZE)}: Size of household: 1, 2, 3, 4, 5, 6, or 7 or more

\item \textbf{Household type (HHT)}: One of 7 types: married couple household, other family household (with a male/female householder), nonfamily household (with a male/female householder, living alone/not living alone).

\item \textbf{Household type; detailed (HHT2)}: One of 12 types: married couple (with/without children $<$ 18), cohabiting couple (with/without children $<$ 18), no spouse/partner present (male/female householder, with own children $<$ 18/with relatives and without own children $<18$/only nonrelatives present/living alone)

\item \textbf{Hispanic householder status (THHSPAN)}: Whether or not the householder is of Hispanic origin,

\item \textbf{Householder age (THHLDRAGE)}: Age of the householder in one of 7 age buckets: 15-24, 24-35, ..., 75-84, or 85 years and older. 

\item \textbf{Householder race (THHRACE)}: Race of the householder in one of 7 categories: White alone, Black alone, American Indian or Alaskan Native alone, Asian alone, Native Hawaiian or Pacific Islander alone, some other race alone, or two or more races.

\item \textbf{Presence of people under 18 years in household (TP18)}: Whether or not one or more people younger than 18 are in the household

\item \textbf{Presence of people over 65 years in household (TP65)}: Whether or not one or more people 65 years and over are in the household

\end{itemize}

\noindent\textbf{Statistics.} We list the household-level Summary File 1 tables names below, along with the descriptors given by the Census Bureau.
\begin{itemize}[noitemsep, leftmargin=10pt]
  \renewcommand{\labelitemi}{}

\item \textbf{P16}: Household type

\item \textbf{P16 A-G}: Household type (iterated by race)

\item \textbf{P16 H}: Household type for households with a householder who is Hispanic or Latino

\item \textbf{P16 I-O}: Household type for households with a householder who is not Hispanic or Latino (iterated by race)

\item \textbf{P16 P-V}: Household type for households with a householder who is Hispanic or Latino (iterated by race)

\item \textbf{P19}: Households by presence of people 65 years and over, household size, and household type

\item \textbf{P20}: Households by type and presence of own children under 18 years

\item \textbf{P21}: Households by presence of people under 18 years

\item \textbf{H1}: Housing units (total count)

\item \textbf{H3}: Occupancy status

\item \textbf{H4}: Tenure

\item \textbf{H4 A-G}: Tenure (iterated by race)

\item \textbf{H4 H}: Tenure of housing units with a householder who is Hispanic or Latino

\item \textbf{H4 I-O}: Tenure of housing units with a householder who is not Hispanic or Latino (iterated by race)

\item \textbf{H4 P-V}: Tenure of housing units with a householder who is Hispanic or Latino (iterated by race)

\item \textbf{H5}: Vacancy status of vacant housing units

\item \textbf{H6}: Race of householder

\item \textbf{H7}: Hispanic or Latino origin of householder by race of householder

\item \textbf{H9}: Household size

\item \textbf{H10}: Tenure by race of householder

\item \textbf{H11}: Tenure by Hispanic or Latino origin of Householder

\item \textbf{H12}: Tenure by household size

\item \textbf{H12 A-G}: Tenure by household size (iterated by race)

\item \textbf{H12 H}: Tenure by household size of households with a householder who is Hispanic or Latino

\item \textbf{H12 I}: Tenure by household size of households with a householder who is White only and not Hispanic or Latino

\item \textbf{H13}: Tenure by age of householder

\item \textbf{H13 A-G}: Tenure by age of householder (iterated by race)

\item \textbf{H13 H}: Tenure by age of householder for housing units with a householder who is Hispanic or Latino

\item \textbf{H13 I}: Tenure by age of householder for housing units with a householder who is White alone and not Hispanic or Latino

\item \textbf{H14}: Tenure by household type by age of householder

\item \textbf{H15}: Tenure by presence of people under 18 years, excluding householders, spouses, and unmarried partners

\end{itemize}

\noindent\textbf{IP Solver.} We use the Gurobi Optimizer to solve our integer programming optimization problems. We specify parameters ``feasibility tolerance'' and ``integer feasibility tolerance'' to their smallest value of $10^{-9}$ to enforce constraints as tightly as possible. We set ``pool search mode'' to 2 in order to find as many solutions as possible, up to the some maximum number defined by ``pool solutions'', which is set to $K=100$ when generating claims and to $1$ when validating claims. We also set the ``timeout'' parameter to 3 minutes to control the total runtime of our experiments for the validation step. In cases where Gurobi times out, we mark that claim as unverified.

\subsection{Additional results for verifying \textit{all} claims}\label{appx:additional_results}

In the main body of our work, we focus on ``singling out'' (singleton claims). However, we note that data reconstruction for multiplicity $m > 1$ can be equally interesting (or privacy violating).
Thus, we present in Figure \ref{fig:results} and Table \ref{tab:results} results for all claims (not just singletons). In general, we make conclusions similar to those in Section \ref{sec:experiments}. The baseline probabilities of most claims are still extremely small, and as expected, more claims (about more households) can be made. For example, Table \ref{tab:results} shows that now, approximately a quarter of households are identified by claims of $k=8$ columns and 70\% by claims of $k=6$ columns.

% \clearpage

% \begin{table*}[t!]
%     \centering
%     \caption{
%         Query Statistics
%     }
%     \begin{tabular}{lrrrrrrrrrrr}
%     \toprule
%     Count ($q(x)$) & 0 & 1 & 2 & 3 & 4 & 5 & 6 & 7 & 8 & 9 & 10 \\
%     \midrule
%     Average \# Queries (out of 621) & 564.39 & 21.12 & 10.38 & 6.04 & 4.44 & 2.87 & 2.63 & 2.01 & 1.82 & 1.74 & 3.57 \\
%     Average Percentage (\%) & 90.88 & 3.40 & 1.67 & 0.97 & 0.71 & 0.46 & 0.42 & 0.32 & 0.29 & 0.28 & 0.58 \\
%     \bottomrule
%     \end{tabular}
%     \label{tab:query_stats}
% \end{table*}

\end{document}